\documentclass[10pt,twocolumn,letterpaper]{article}

\usepackage{cvpr}
\usepackage{times}
\usepackage{epsfig}
\usepackage{graphicx}
\usepackage{amsmath}
\usepackage{amssymb}
\usepackage{xcolor}
\usepackage[export]{adjustbox}

\usepackage[english]{babel}

\newcommand{\placetextbox}[3]{
	\setbox0=\hbox{#3}
	\AddToShipoutPicture*{
		\put(\LenToUnit{#1\paperwidth},\LenToUnit{#2\paperheight}){\vtop{{\null}\makebox[0pt][c]{#3}}}%
	}%
}%

\usepackage[pagebackref=true,breaklinks=true,letterpaper=true,colorlinks,bookmarks=false]{hyperref}

\cvprfinalcopy 

\def\cvprPaperID{11} 

\ifcvprfinal\fi

\definecolor{todo}{rgb}{7,.0,.3} 

\begin{document}
	
\placetextbox{0.5}{0.96}{\fbox{
		\begin{tabular}{@{}c@{}}
		
		\footnotesize
		\textcopyright2019 IEEE. Personal use of this material is permitted.  Permission from IEEE must be obtained for all other uses, in any current or future media, including \\
		\footnotesize  reprinting/republishing this material for advertising or promotional purposes, creating new collective works, for resale or redistribution to servers or lists, \\
		\footnotesize  or reuse of any copyrighted component of this work in other works.

		\end{tabular}	
}}%

\title{Leveraging Semantic Embeddings for Safety-Critical Applications}


\author{
\begin{tabular}[t]{c}
Thomas Brunner\textsuperscript{1,2}
\and
Frederik Diehl\textsuperscript{1,2}
\and
Michael Truong Le\textsuperscript{1,2}
\and
Alois Knoll\textsuperscript{2}
\end{tabular} \\
\begin{tabular}[t]{c}
\textsuperscript{1} fortiss GmbH
\and
\textsuperscript{2} Technical University of Munich
\end{tabular}\\
\begin{tabular}[t]{c}
{\tt\small \{brunner,diehl,truongle\}@fortiss.org}
\and
{\tt\small knoll@in.tum.de}
\end{tabular}
}

\maketitle
\thispagestyle{empty}

\begin{abstract}
	Semantic Embeddings are a popular way to represent knowledge in the field of zero-shot learning. We observe their interpretability and discuss their potential utility in a safety-critical context. Concretely, we propose to use them to add introspection and error detection capabilities to neural network classifiers. First, we show how to create embeddings from symbolic domain knowledge. We discuss how to use them for interpreting mispredictions and propose a simple error detection scheme. We then introduce the concept of semantic distance: a real-valued score that measures confidence in the semantic space. We evaluate this score on a traffic sign classifier and find that it achieves near state-of-the-art performance, while being significantly faster to compute than other confidence scores. Our approach requires no changes to the original network and is thus applicable to any task for which domain knowledge is available.
\end{abstract}


\section{Introduction}

Despite their remarkable performance, deep neural networks often produce errors (i.e. mispredictions) that seem illogical to a human observer. Why was a traffic sign misclassified? Why was a pedestrian not detected? What was the internal state of the network at the time, and what information was contained?
 
 Naturally, these questions are of great interest when developing safety-critical applications. Consider the field of automated driving: In an industry that depends not only on safety, but also on its perception by customers, there is a need for systems that can explain their decisions, and do so in a way that looks rational to humans.
 
But this is currently not the case. In large neural networks, knowledge is typically so entangled that it cannot be easily interpreted \cite{Bengio:2013:RLR:2498740.2498889}. Even worse, when a mistake is made, predictors often report high confidence scores (e.g. through softmax activations), when in reality they should report significant uncertainty \cite{Gal:2016:DBA:3045390.3045502}.

 
 


So what is the missing link? Humans are often equipped with additional domain knowledge that captures the semantics of the task at hand. This allows them to judge whether a result seems plausible and to discard it otherwise. It would be desirable to have systems with neural networks do the same: capture semantics of the current situation, use this knowledge to perform sanity checks and finally report the confidence they have in their own decisions.
 
 To achieve this goal, we draw inspiration from the field of zero-shot learning. There, a variety of methods exists for constructing so-called semantic embeddings \cite{Palatucci2009}, which encode semantic information into vector spaces and can easily be applied to neural network features.
 
 In zero-shot learning, these projections are used to recognize images of previously unseen classes, based on their semantic attributes. Here, we propose to leverage the same representations in a safety context, and thereby gain interpretability and error detection capabilities.

 \begin{figure}[tbp]

 	\centering
 	\includegraphics[width=.46\textwidth]{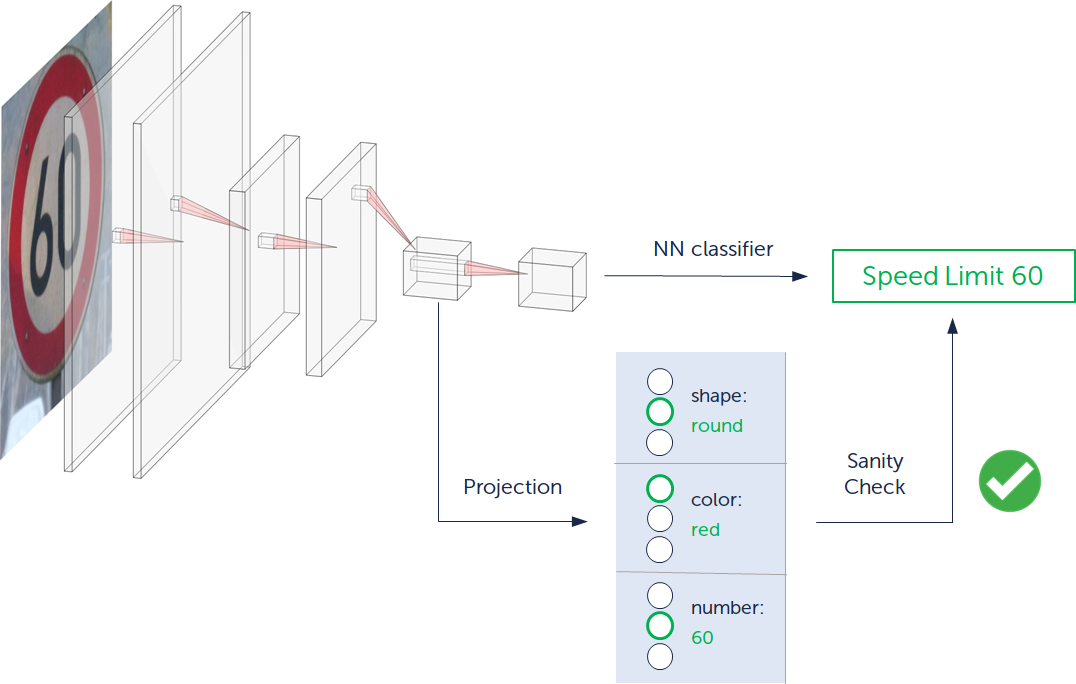}
 	
 	\caption{Semantic embedding for a traffic sign classifier. Features are projected to a representation which is directly derived from domain knowledge about the classification task.}
 	\label{fig:teaser}	
 \end{figure}
 \
 \\
 \
Our contribution is as follows:
 \begin{itemize}
	 \item We discuss how to form semantic embeddings from domain knowledge and how to use them to perform sanity checks on neural network predictions.
	 \item We generalize this notion to a real-valued \textit{semantic confidence} score, which can be used like an uncertainty estimate.
	 \item In a proof of concept, we show that this simple score achieves state-of-the-art performance on a selective classification task, while at the same time being significantly faster than established methods.
 \end{itemize}

\section{Related work} \label{sec:related_work}

\subsection{Leveraging symbolic knowledge}

The idea of combining expert domain knowledge with machine learning techniques has a long history. In the 1990s, much work went into the formulation of so-called hybrid learning methods which tried to directly inject symbolic knowledge into machine learning models \cite{Towell:1994:KAN:194414.194434}, or to extract rules from them \cite{d'AvilaGarcez:2001:SKE:362720.362725}.

However, most of these techniques are impractical to use with the neural networks of today, which are very large and trained with great amounts of data. For example, a symbolic initialization as proposed in \cite{Towell:1994:KAN:194414.194434} would certainly be overwritten by the lengthy training procedure.
Similarly, symbolic rule extraction from neural networks \cite{d'AvilaGarcez:2001:SKE:362720.362725} is computationally prohibitive for large networks, and the extracted knowledge is hardly interpretable by humans. Hence, those techniques are difficult to apply to the networks of today.

\subsection{Semantic embeddings}

More recently, another way to leverage domain knowledge has emerged in the field of zero-shot learning: semantic embeddings. In their simplest form, they can be described as feature spaces where each dimension encodes a high-level semantic property \cite{Palatucci2009}. 

The zero-shot learning task consists of recognizing examples of previously unseen classes, and many techniques exist to create suitable embeddings for it. Perhaps the simplest is Direct Attribute Prediction (DAP) \cite{Lampert2009}, where a classifier learns to identify semantic attributes that are defined in a symbolic knowledge base. The class can then be inferred by reasoning over these attributes. 

There are more sophisticated forms of embeddings, and most work in this field has moved away from predefined symbols and towards learning representations in a semi-supervised manner \cite{Xian2017ZeroShotL}. However, while this improves performance on the zero-shot learning task, it also destroys two major aspects that we are interested in: symbolic interpretability and the ability to import those symbols directly from an existing knowledge base. For this reason, we focus on the most basic technique, which is DAP \cite{Lampert2009}.

In Zero-shot learning, semantic embeddings are used for improved generalization capabilities. But is that the only possible application? It seems obvious that such an embedding space -- especially in the form of human-readable attributes -- could be of tremendous use for interpretability, safety and security. This is precisely our aim: to have a feature space that is human-readable, and to use it to detect classifications that are implausible, or plainly wrong.

\subsection{Measuring confidence}

Typically, implausible predictions are detected by calculating a \textit{confidence score} that increases with the likelihood of a correct classification, and vice versa. By applying a threshold, predictions with low confidence can then be discarded. Here, we describe established methods to calculate such a score before introducing our own in Section \ref{sec:our_method}.

\textbf{Softmax}: Perhaps the simplest way to measure confidence in a neural network classifier is to use the raw softmax output \cite{Cordella1995AMF}. While this method has been shown to produce rather inaccurate confidence scores \cite{Bendale2016TowardsOS,Gal:2016:DBA:3045390.3045502}, it still remains in widespread use to this day.

\textbf{Monte-Carlo Dropout (MCD)} \cite{Gal:2016:DBA:3045390.3045502}: One of the most promising approaches is Monte-Carlo Dropout, which activates Dropout \cite{Srivastava2014DropoutAS} at inference time to approximate the behavior of a Bayesian Neural Network. This method samples the network multiple times and produces a distribution of outputs, from which uncertainty can then be estimated. This method can be applied without modification to the network, but requires training with Dropout. It has been shown to perform well, but is very expensive to compute: multiple forward passes must be performed (Gal et al. \cite{Gal:2016:DBA:3045390.3045502} recommend 100) to produce the estimate.

\textbf{NN-Distance (NND)} \cite{Mandelbaum2017DistancebasedCS}: This approach is perhaps the closest to our own work. The authors propose to use the feature space of the penultimate layer to calculate the Euclidean distance between an example and its nearest neighbors in the training set. Intuitively, if this distance is high, then the example is an outlier and the prediction is less confident. They report high performance, but only if the network is retrained with an additional loss term specifically geared towards their score. While NND only needs one forward pass, it has to store a representation of the training data and perform k-Nearest-Neighbor search on it. With various optimizations, time and memory requirements are much lower than for MCD, but still higher than for our method, as we discuss in Section \ref{sec_eval}.



\section{Semantic embeddings for safety} \label{sec:our_method}

We now present our approach for detecting mispredictions. We first discuss how to construct semantic embeddings from a symbolic knowledge base, then describe how to perform simple error detection, and finally introduce the concept of \textit{semantic distance}, which is a general-purpose confidence score that is directly derived from the provided domain knowledge.

\subsection{Design of the embedding space}

Let $F = \{f_1, \ldots, f_m\} \in \mathbb{R}^{m \times n}$ denote the features of a data set, where $m$ is the number of examples and $n$ the number of features provided by a feature extractor (e.g. the last hidden layer of a neural network). Then $S = \{s_1, \ldots, s_m\} \in \mathbb{R}^{m \times k}$ is the corresponding semantic embedding of the examples, where each example is represented by a $k$-dimensional attribute vector. We refer to this as the \textit{semantic space}.

How does one know which attributes to choose for this space? Technically any form of embedding can be used, but for the sake of simplicity we limit ourselves to DAP, with attributes generated from a knowledge base $K$. Formally, given the class labels $Y$, $K$ provides a mapping $Y \rightarrow S$ that can be used to generate semantic annotations $S_Y$ from the ground truth of an annotated data set.



The simplest way to design such a knowledge base is to manually define  attributes that intuitively make sense to a human, so that they can later be used to explain the predicted label. For example, if our goal is to create an embedding for traffic sign classification, we might define attribute groups such as color, shape, pictogram, etc., one-hot encode them, and then concatenate everything into a single vector (see Figure \ref{fig:teaser}). The goal is not to achieve a perfect layout of the embedding space, but to have it contain attributes that intuitively make sense to a human. Section \ref{sec_eval} contains a description of the knowledge base that we use in our experiments.



\subsection{Projection method}

Similar to the design of the embedding space, the choice of projection method is not critical for showcasing our approach. We therefore use the Semantic Autoencoder (SAE) \cite{Kodirov2017SemanticAE} technique, which is applied to the last hidden layer of the network and learns a linear projection from it into the attribute space. This method is simple, fast and has recently shown remarkable performance on the Zero-Shot Learning task.

SAE learns a projection matrix $W$, so that $S = W \times F$. As an additional constraint, the same matrix is used to project back into feature space: $\hat{F} = W^T \times S$. $W$ is then obtained by minimizing the reconstruction error $\lvert\lvert F - \hat{F} \lvert\lvert_2$. The resulting matrix $W$ is essentially a linear classifier, but the added constraint has been shown to drastically improve the quality of the semantic representation \cite{Kodirov2017SemanticAE}.

To create the embedding for an existing classifier, one can simply reuse the original training data and extract features and semantic annotations from it. Now that both $F_{train}$ and $S_{Y_{train}}$ are provided, $W$ can be efficiently calculated with the Bartels-Stewart algorithm. This is very fast and, as we show in Section \ref{sec_eval}, can be applied to large data sets in a matter of seconds.


 \begin{figure}[tbpb]
 	
 	\centering
 	\includegraphics[width=.23\textwidth]{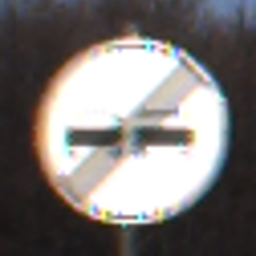}
 	\includegraphics[width=.23\textwidth]{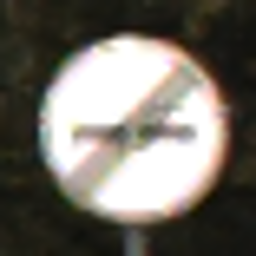}
 	
 	\caption{Examples of mispredictions. The classifier wrongly predicts "no overtaking" instead of "end of no overtaking". However, our embedding shows that the features recognized by the network are not consistent with the knowledge base: after projection, the semantic vector for this prediction states "round, red, crossed out, cars, no number", which is not a valid configuration. This is detected as an error, and the prediction discarded. The color attribute in the semantic vector also hints at the likely cause: Chromatic aberration may have excited neurons that respond to red.}
 	\label{fig:signs}	
 \end{figure}

\subsection{Detecting errors}

For any input, given the feature vector $f$ and projection $W$, we receive the attribute vector $s_{pred} = Wf$. We also have a set of valid configurations $S_{proto} = \{s_1, \ldots, s_c\}$, each of which is the prototype of a class label in the semantic space (provided by the knowledge base, with $c$ being the number of classes). To check for an error, we merely need to compare $s_{pred}$ to all elements in $S_{proto}$. 

The concept of prototypes is certainly not new. Typically, they are used for classification tasks and are created directly from the input (training) data \cite{Snell2017PrototypicalNF}. Our prototypes however are not created from examples, but purely from knowledge about the classes. In this way, we invert the idea: our semantic space is an embedding of labels, not of examples. 

The easiest way to detect an error is to use $argmax$ on each individual attribute group and then do a binary comparison. See Figure \ref{fig:signs}, where an invalid prediction is recognized, as the example $s_{pred} \notin S_{proto}$. This is a strong indicator that the feature extractor has performed poorly, and therefore the prediction given by the original classifier is unlikely to be correct.

In this way, the semantic embedding forms an error-detecting code with human-readable explanations. This in itself is already useful, as it performs a much-desired "sanity check", but it also provides a lead as to where the feature extractor might have a specific weakness.

\subsection{Semantic distance as a confidence score}

Binarizing the attributes is useful, but it incurs a great loss of information. Instead of clamping the values of $s_{pred}$, hoping to hit a valid attribute configuration, we can simply measure the distance to the prototype of the predicted class in the embedding space.

Let $y$ be the class label originally predicted by the classifier. Then $s_{y} \in S_{proto}$ is the prototype attribute configuration for this class, as obtained from the knowledge base. We can now compare $s_{pred}$ and $s_{y}$ with a distance metric of our choice. Since the attribute vectors are not normalized, we use cosine similarity to formulate the following \textit{semantic distance} score:

\[
d=1-\cos(\theta) = 1 - \dfrac{s_{pred} \circ s_{y}}{\lvert\lvert s_{pred}\lvert\lvert \  \lvert\lvert s_{y}\lvert\lvert }
\]

Hence, $d$ is the rotational error between a prediction and the class label in the semantic space. This allows us to use $d$ as a general-purpose confidence score: 0 is a perfect match and 1 indicates orthogonality. By introducing a threshold $\epsilon$, we can formulate a selective classification system that finds and discards predictions with high semantic distance, i.e. for which $d > \epsilon$. Now we can directly evaluate our method against existing confidence scores.


\section{Experiments}
\label{sec_eval}

We apply our method to a neural network classifier on the GTSRB data set \cite{GTSRB}. First, we show an example of how typical mispredictions can be detected, discarded and explained with semantic attributes. 
We then proceed to benchmark our semantic distance score against two recently proposed confidence scores, Monte-Carlo Dropout (MCD) and NN-Distance (NND), on the task of selective classification.

\subsection{Setup}
\textbf{Data set}: Our method can be applied to any classifier, but we cannot choose an arbitrary data set. In its current form, our approach depends on domain knowledge which is provided by a knowledge base of semantic attributes. It would be difficult to formulate sensible attributes for classes of MNIST or CIFAR, on which most confidence scores have been benchmarked so far. This is a drawback which we aim to address in future work (see Section \ref{sec:conclusion}).
 
 Instead, we choose to evaluate our method on the \textit{German Traffic Sign Recognition Benchmark} (GTSRB) \cite{GTSRB}. This data seems to be a natural fit: Semantic attributes can be easily formulated, and traffic sign classification is a good example of a safety-critical application. The GTSRB has 43 different classes and consists of 39209 training and 12630 test images.

\textbf{Knowledge base}: In order to describe the semantic attributes of a broad range of traffic signs, we create a knowledge base of five multi-valued attribute groups: 
 \begin{itemize}
 	\item Shape (5): round, triangular, ...
 	\item Color (4): red, blue, ...
 	\item Crossed out (2): yes, no
 	\item Pictogram (29): none, number, children, frost, ...
 	\item Number (9): none, 20, 30, ...
 \end{itemize}

Each traffic sign has exactly one value for each attribute group, i.e.  values are mutually exclusive and the zero vector is avoided. In this way, a "speed limit 60" sign is defined by "round, red, not crossed out, number, 60". The resulting semantic space is of dimensionality $k=49$. 
 
\textbf{Classifier}: As the GTSRB benchmark is a rather simple task, we can use a reasonably small model to demonstrate our concept. Our network consists of 3 blocks of 2 convolutional layers each, one fully-connected layer and an output layer with softmax. We train with Dropout after each block and after the fully-connected layer. After training, the classifier achieves 99.27\% accuracy on the test set, i.e. there are roughly 100 mispredictions. 

\textbf{Semantic Embedding}: We apply SAE to the last hidden layer of the classifier. Using the knowledge base, we generate ground truth semantic annotations for all examples, and then optimize the SAE equation over the entire training set to obtain the projection. SAE has a regularizing parameter $\lambda$, for which we have found $0.1$ to work well.

\subsection{Interpreting mispredictions}

To display the individual attributes of a prediction, we can simply split the predicted semantic vector $s_{pred}$ into its individual attribute groups and perform $argmax$ on each group. This can be used for error detection, but it may also give us valuable hints about what may have been the cause for misprediction.

Consider Figure \ref{fig:signs}: we curiously find that the classifier sometimes confuses "no overtaking" with "end of no overtaking" signs. A glance at the semantic vector reveals the likely cause: for both examples, the semantic vector shows "red" as the dominant attribute in the "color" group. This means that, somewhere inside the classifier, those neurons that are trained to react to red traffic signs have high activations. And indeed, we find that the images feature red chromatic aberration, an effect that seems to be mostly absent from the training set. 

In this way, the embedding has helped us identify a weakness in the classifier that could be remedied, for example by implementing data augmentation strategies. In any case, since the semantic vector does not correspond to a valid class prototype, the error is detected and the prediction can be safely discarded.

\subsection{Confidence scores}

We now benchmark our proposed semantic distance score against MCD and NND on the task of selective classification. 

\textbf{MCD}: \cite{Gal:2016:DBA:3045390.3045502}: Since our network is trained with Dropout, we can simply activate it at inference time. For each example in the test set, we run 100 forward passes and estimate uncertainty from the output variance. 

\textbf{NND} \cite{Mandelbaum2017DistancebasedCS}: We extract the features of the penultimate layer for the entire training set. At inference time, we calculate NND with all training examples. The authors recommend various schemes to retrain the classifier in order to increase the performance of their method, but we refrain from this, as we are strictly interested in a plug-and-play scenario, where no modification is necessary.

\textbf{Selective classification}: Distance scores and uncertainty estimates generally approach zero as confidence increases. Therefore, we can define a threshold $\epsilon$, above which an example is rejected. In other words, this is a detection task where every correct prediction should have high confidence, and mispredictions should be detected via the threshold.

As is commonly done for detectors, we use ROC curves that balance true positives against false positives at a moving $\epsilon$. We then use the area under this curve (AUC, higher is better) to directly compare all methods.

 \begin{figure}[tbp]

 	\centering
 	\includegraphics[width=.49\textwidth]{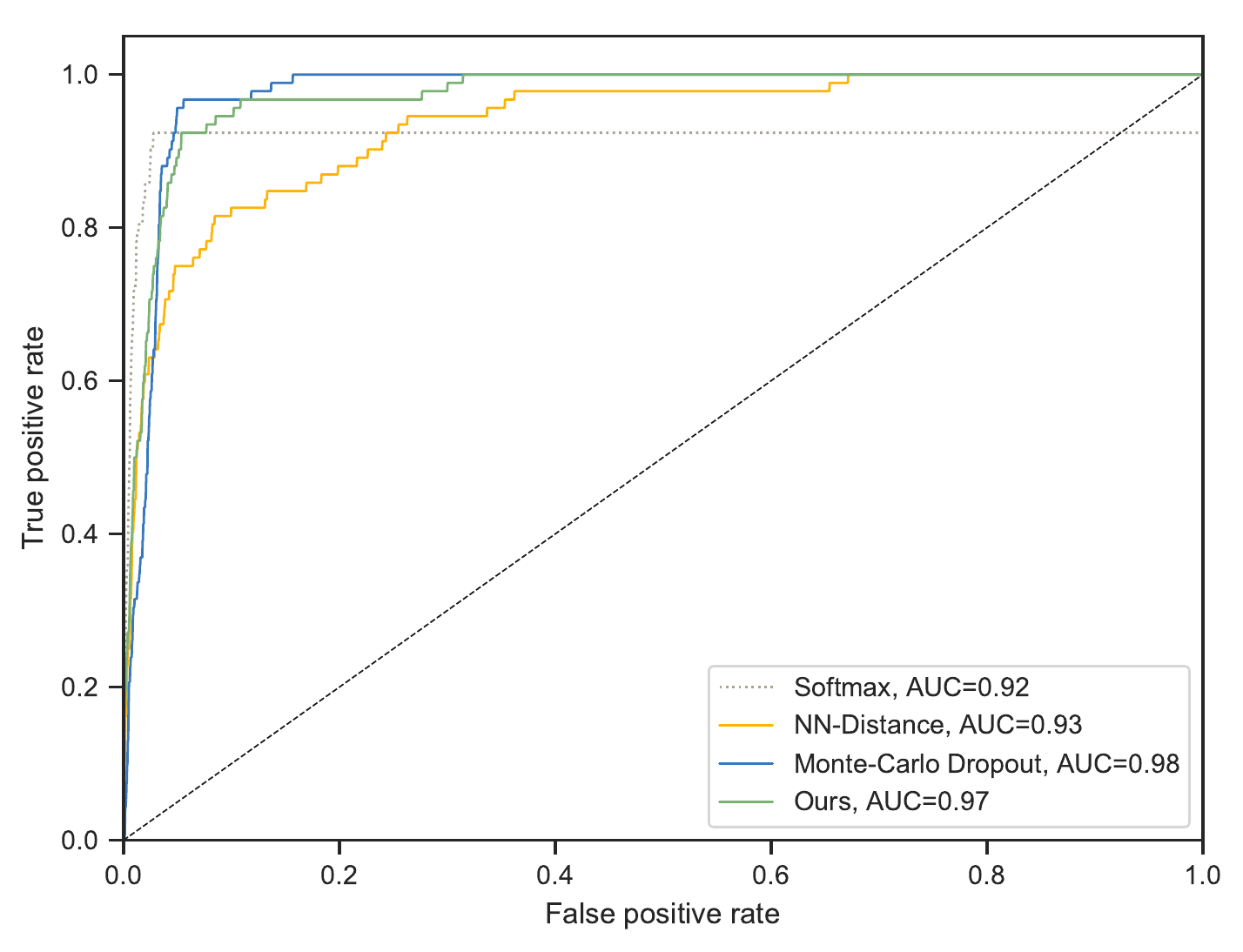}
 	
 	\caption{ROC curves for several confidence scores. The goal is to achieve a high true positive rate, while at the same time keeping false positives low. Our proposed semantic distance score is similar in performance to MCD, while requiring only one forward pass to compute. Softmax fails to reach 1.0 as it reports the maximum confidence value for a range of misclassifications.}
 	\label{fig:roc}	
 \end{figure}

\textbf{Results}: Figure \ref{fig:roc} shows the result of our evaluation. The most effective method is MCD and our proposed semantic distance is in second place, having only slightly lower AUC. This means that we do not outperform the state of the art, but almost match it with a method that is both more interpretable and significantly faster. 

NND exhibits slightly lower performance. We are aware that the authors suggest various strategies to retrain the classifier before performing detection, and that the setting we consider does not allow this -- therefore our experiment does not play to their strengths. Nevertheless, it shows that simply performing out-of-distribution detection in the feature space of a hidden layer may not be an optimal strategy when used out-of-the-box. We do note however that retraining a network to improve the layout of a feature space is conceptually close to engineering an embedding. This connection seems interesting, and we aim to investigate it in future work.

Finally, Figure \ref{fig:roc} also shows a confidence score obtained from Softmax. It performs very well at first but then fails to detect the last remaining mispredictions. This is in line with prior work -- Softmax has a tendency to report perfect scores (1.0) for classifications in which the model actually has high uncertainty \cite{Bendale2016TowardsOS,Gal:2016:DBA:3045390.3045502}.

\textbf{Runtime considerations}: The strongest method, MCD, is also the slowest. Running 100 forward passes for a single prediction may incur prohibitive costs for many applications. NND requires only one forward pass, but needs to compare the features of an input against a large number of data points in the training set (the authors recommend all of them). K-nearest-neighbor search can be heavily optimized, but still requires significantly more calculation than our method: The semantic distance can be obtained by performing a single dot product with the prototype of the predicted class.

\section{Conclusion}
\label{sec:conclusion}

We have shown how Semantic Embeddings can be applied to Neural Networks for the purpose of interpretability and safety. These representations are generated from expert domain knowledge, offer insight into the nature of mispredictions and allow for automated detection of errors. We have also proposed a new confidence score that measures \textit{semantic distance} in the space of such an embedding and that can be used in much of the same way as an uncertainty estimate.

In a simple proof of concept, we have shown that this score achieves very good performance on a selective classification task, while being significantly easier to compute than other confidence measures. Our score does not require modification of the original network and is thus easily implemented whenever a knowledge base of semantic attributes is available.

Our approach has one drawback: in its current form, it requires a manually defined knowledge base. The problem could be approached in two ways: symbolic knowledge could be acquired and transformed from existing large-scale knowledge bases (e.g. CYC). Alternatively, one could automatically learn the semantics of the embedding from data -- this is also the direction on which the field of zero-shot learning is currently focused. We expect this to improve performance, but at the same time degrade interpretability. 
It should be interesting to investigate closer the nature of this trade-off, and to identify the deciding factors that lead to good embeddings, strong interpretability and high-quality confidence scores. 

Overall, we have shown that semantic embeddings can be applied to safety-critical applications in a way that is both simple and powerful, and we believe that this combination holds much promise for the future.

\section*{Acknowledgements}

This research has been supported by the Bavarian Ministry of Economic Affairs, Regional Development and Energy as part of the fortiss project "Dependable AI".

{\small
\bibliographystyle{ieee}
\bibliography{egbib}

\begin{thebibliography}{10}\itemsep=-1pt

\bibitem{Bendale2016TowardsOS}
A.~Bendale and T.~E. Boult.
\newblock Towards open set deep networks.
\newblock In {\em Proceedings of the 2016 IEEE Conference on Computer Vision
  and Pattern Recognition}, CVPR'16, pages 1563--1572, 2016.

\bibitem{Bengio:2013:RLR:2498740.2498889}
Y.~Bengio, A.~Courville, and P.~Vincent.
\newblock Representation learning: A review and new perspectives.
\newblock {\em IEEE Trans. Pattern Anal. Mach. Intell.}, 35(8):1798--1828, Aug.
  2013.

\bibitem{Cordella1995AMF}
L.~P. {Cordella}, C.~{De Stefano}, F.~{Tortorella}, and M.~{Vento}.
\newblock A method for improving classification reliability of multilayer
  perceptrons.
\newblock {\em IEEE Transactions on Neural Networks}, 6(5):1140--1147, 1995.

\bibitem{d'AvilaGarcez:2001:SKE:362720.362725}
A.~S. d'Avila Garcez, K.~Broda, and D.~M. Gabbay.
\newblock Symbolic knowledge extraction from trained neural networks: A sound
  approach.
\newblock {\em Artif. Intell.}, 125(1-2):155--207, Jan. 2001.

\bibitem{Gal:2016:DBA:3045390.3045502}
Y.~Gal and Z.~Ghahramani.
\newblock Dropout as a bayesian approximation: Representing model uncertainty
  in deep learning.
\newblock In {\em Proceedings of the 33rd International Conference on Machine
  Learning}, ICML'16, pages 1050--1059, 2016.

\bibitem{Kodirov2017SemanticAE}
E.~Kodirov, T.~Xiang, and S.~Gong.
\newblock Semantic autoencoder for zero-shot learning.
\newblock In {\em Proceedings of the 2017 IEEE Conference on Computer Vision
  and Pattern Recognition}, CVPR'17, pages 4447--4456, 2017.

\bibitem{Lampert2009}
C.~Lampert, H.~Nickisch, and S.~Harmeling.
\newblock Learning to detect unseen object classes by between-class attribute
  transfer.
\newblock In {\em Proceedings of the 2009 IEEE Conference on Computer Vision
  and Pattern Recognition}, CVPR'09, pages 951--958, 2009.

\bibitem{Mandelbaum2017DistancebasedCS}
A.~Mandelbaum and D.~Weinshall.
\newblock Distance-based confidence score for neural network classifiers.
\newblock {\em arXiv preprint arXiv:1709.09844}, 2017.

\bibitem{Palatucci2009}
M.~Palatucci, D.~Pomerleau, G.~Hinton, and T.~M. Mitchell.
\newblock Zero-shot learning with semantic output codes.
\newblock In {\em Proceedings of the 22nd International Conference on Neural
  Information Processing Systems}, NIPS'09, pages 1410--1418, 2009.

\bibitem{Snell2017PrototypicalNF}
J.~Snell, K.~Swersky, and R.~S. Zemel.
\newblock Prototypical networks for few-shot learning.
\newblock In {\em Neural Information Processing Systems}, NIPS'17, 2017.

\bibitem{Srivastava2014DropoutAS}
N.~Srivastava, G.~E. Hinton, A.~Krizhevsky, I.~Sutskever, and R.~R.
  Salakhutdinov.
\newblock Dropout: a simple way to prevent neural networks from overfitting.
\newblock {\em Journal of Machine Learning Research}, 15:1929--1958, 2014.

\bibitem{GTSRB}
J.~Stallkamp, M.~Schlipsing, J.~Salmen, and C.~Igel.
\newblock Man vs. computer: Benchmarking machine learning algorithms for
  traffic sign recognition.
\newblock {\em Neural networks : the official journal of the International
  Neural Network Society}, 32:323--32, 02 2012.

\bibitem{Towell:1994:KAN:194414.194434}
G.~G. Towell and J.~W. Shavlik.
\newblock Knowledge-based artificial neural networks.
\newblock {\em Artif. Intell.}, 70(1-2):119--165, Oct. 1994.

\bibitem{Xian2017ZeroShotL}
Y.~Xian, C.~H.~Lampert, B.~Schiele, and Z.~Akata.
\newblock Zero-shot learning - a comprehensive evaluation of the good, the bad
  and the ugly.
\newblock {\em IEEE Transactions on Pattern Analysis and Machine Intelligence},
  PP, 2017.

\end{thebibliography}
}

\end{document}